\documentclass[letterpaper]{article}
\usepackage{uai2019}
\usepackage[margin=1in]{geometry}

\usepackage{times}

\usepackage[utf8]{inputenc} 
\usepackage[T1]{fontenc}    
\usepackage{hyperref}       
\usepackage{url}            
\usepackage{booktabs}       
\usepackage{amsfonts}       
\usepackage{nicefrac}       
\usepackage{microtype}      

\usepackage{booktabs}
\usepackage{amsmath}
\usepackage{mathrsfs}
\usepackage{amsthm}
\usepackage{stmaryrd}
\usepackage{amssymb}

\usepackage{relsize}
\usepackage{pifont}
\usepackage{multirow}
\usepackage{color}
\usepackage{mathtools}
\usepackage{tikz}
\usetikzlibrary{arrows}
\usepackage{algorithm}
\usepackage{todonotes}
\usepackage{wrapfig}
\usepackage{caption}

\usepackage[round]{natbib}
\usepackage{multibib} %

\usepackage{graphicx}

\renewcommand{\S}{\mathcal{S}}
\newcommand{\A}{\mathcal{A}}

\newcommand{\Real}{\mathbb{R}}

\newcommand{\E}{\mathbb{E}}
\newcommand{\EE}[1]{\E\left[#1\right]}
\newcommand{\cbar}{\,|\,}

\usepackage{algpseudocode,algorithm,algorithmicx}

\hypersetup{
	plainpages=false,
	colorlinks=true,              
	linkcolor=blue,               
	anchorcolor=blue,             
	citecolor=blue,               
	filecolor=blue,               
	pagecolor=blue,               
	urlcolor=blue,                
	pdfview=FitH,                 
	pdfstartview=FitH,            
	pdfpagelayout=SinglePage      
}

\title{Randomized Value Functions via Multiplicative Normalizing Flows}

\author{} 

\author{Ahmed Touati\textsuperscript{1, 3}\ \ Harsh Satija\textsuperscript{2, 3}\ \ Joshua Romoff\textsuperscript{2, 3}\ \ Joelle Pineau \textsuperscript{2, 3}\ \  Pascal Vincent\textsuperscript{1, 3} \\ 
  \textsuperscript{1} Mila, Universit\'e de Montr\'eal\ \ \ \ \
  \textsuperscript{2}Mila, McGill University \ \ \ \ \
  \textsuperscript{3}Facebook AI Research
  }

\begin{document}

\maketitle

\begin{abstract}
Randomized value functions offer a promising approach towards the challenge of efficient exploration in complex environments with high dimensional state and action spaces. Unlike traditional point estimate methods, randomized value functions maintain a posterior distribution over action-space values. 
This prevents the agent's behavior policy from prematurely exploiting early estimates and falling into local optima.
In this work, we leverage recent advances in variational Bayesian neural networks and combine these with traditional Deep Q-Networks (DQN) and Deep Deterministic Policy Gradient (DDPG) to achieve randomized value functions for high-dimensional domains. In particular, we augment DQN and DDPG with multiplicative normalizing flows in order to track a rich approximate posterior distribution over the parameters of the value function. 
This allows the agent to perform approximate Thompson sampling in a computationally efficient manner via stochastic gradient methods. We demonstrate the benefits of our approach through an empirical comparison in high dimensional environments.
\end{abstract}

\section{INTRODUCTION}
Efficient exploration is one of the main obstacles in scaling up modern deep reinforcement learning (RL) algorithms \citep{bellemare2016unifying, osband2017deep, fortunato2017noisy}. The main challenge in efficient exploration is the balance between exploiting current estimates, and gaining information about poorly understood states and actions.
Despite the wealth of research into provably efficient exploration strategies, most of these 
focus on tabular representations and are typically intractable in high dimensional environments \citep{strehl2005theoretical, kearns2002near, brafman2002r}. Currently, the most widely used technique, in Deep RL, involves perturbing the greedy action with some local random noise, e.g $\epsilon$-greedy or Bolzmann exploration \citep{sutton1998introduction}. This naive perturbation is not directed; it continuously explores actions that are known to be sub-optimal and may result in sample complexity that grows exponentially with the number of states \citep{kearns2002near, osband2017deep}.

\textit{Optimism in the face of uncertainty} is one of the traditional guiding principles that offers provably efficient learning algorithms \citep{strehl2005theoretical, kearns2002near, brafman2002r, jaksch2010near}. These algorithms incentivize learning about the environment by rewarding the discoveries of poorly understood states and actions with an exploration bonus. In these approaches, the agent first builds a confidence set over Markov Decision Processes (MDPs) that contains the true MDP with high probability. Then, the agent determines the most optimistic and statistically plausible version of its model and acts optimally with respect to it. Inspired by this principle, several Deep RL works prescribe guided exploration strategies, such as pseudo-counts \citep{bellemare2016unifying}, variational information maximization \citep{houthooft2016vime} and model prediction errors \citep{stadie2015incentivizing}. All of the aforementioned methods add an intrinsic reward to the original reward and then simply train traditional Deep RL algorithms on the augmented MDP.

An entire body of algorithms for efficient exploration is inspired by \textit{Thompson sampling} \citep{thompson1933likelihood}. \textit{Bayesian dynamic programming} was first introduced in \citet{strens2000bayesian} and 
has become 
known more recently as \textit{posterior sampling for reinforcement learning} (PSRL) \citep{osband2013more}. In PSRL, the agent starts with a prior belief over world models and then proceeds to update its full posterior distribution over models with the newly observed samples. A model hypothesis is then sampled from this distribution, and a greedy policy with respect to the sampled model is followed thereafter. Unfortunately, due to their high computational cost, these methods are only feasible on small MDPs and are of limited practical use in high dimensional environments. 

\citet{osband2017deep} developed \textit{randomized value functions} in order to improve the scalability of PSRL. At an abstract level, randomized value functions can be interpreted as a model-free version of PSRL. Instead of maintaining a posterior belief over possible models, the agent's belief is expressed over value functions. Similarly to PSRL, a value function is sampled at the start of each episode and actions are selected greedily thereafter. Subsequently, actions with highly uncertain values are explored due to the variance in the sampled value functions. In order to scale this approach to large MDPs with linear function approximation,  \citet{osband2016generalization} introduce randomized least-square value iteration (RLSVI) which involves using Bayesian linear regression for learning the value function.

In the present work, we are interested in using randomized value functions with deep neural networks as a function approximator. To address the issues with computational and statistical efficiency, we leverage recent advances in variational Bayesian neural networks. Specifically, we use normalizing multiplicative flows (MNF) \citep{louizos2017multiplicative} in order to account for the uncertainty of estimates for efficient exploration. MNF is a recently introduced family of approximate posteriors for Bayesian neural networks that allows for arbitrary dependencies between neural network parameters. 

Our main contribution is to introduce MNFs into standard value-based Deep RL algorithms, yielding a well-principled and powerful method for efficient exploration. 
We validate 
this approach experimentally by comparing against recent Deep RL baselines on several challenging exploration domains, including the Arcade Learning Environment (ALE) \citep{bellemare2013arcade}. 
We thus show that the richness of the approximate posterior in MNF enables more efficient exploration in deep reinforcement learning.

\section{REINFORCEMENT LEARNING BACKGROUND} 
\label{sec:RL background}

In reinforcement learning, an agent interacts with its environment which is modelled as a discounted Markov Decision Process $(\S, \A, \gamma, P, r)$ with state space $\S$, action space $\A$, discount factor $\gamma \in [0, 1)$, transition probabilities $P : \S \times \A \rightarrow (\S \rightarrow [0,1])$ mapping state-action pairs to distributions over next states, and reward function $r : (\S \times \A) \rightarrow \Real$ ~\citep{sutton1998introduction}. We denote by $\pi(a \cbar s)$ the probability of choosing an action $a$ in the state $s$ under the policy $\pi : \S \rightarrow (\A \rightarrow [0, 1])$. The action-value function for policy $\pi$, denoted $Q^{\pi}:\S \times \A \rightarrow \Real$, represents the expected sum of discounted rewards along the trajectories induced by the MDP and $\pi$: $Q^{\pi}(s, a) = \EE{\sum_{t=0}^{\infty} \gamma^t r_t \cbar (s_0, a_0)=(s, a), \pi}$. The expectation is over the distribution of admissible trajectories $(s_0, a_0, s_1, a_1, \ldots)$ obtained by executing the policy $\pi$ starting from $s_0 = s$ and $a_0 = a$. The action-value function of the optimal policy is $Q^{\star}(s, a) = \arg \max_{\pi}Q^{\pi}(s, a)$ and it satisfies the Bellman optimality equation:
\begin{equation}
Q^\star(s, a) = r(s, a) + \gamma \sum_{s' \in \S}p(s' \cbar s, a) \max_{a \in \A} Q^\star(s', a)
\end{equation}

\textbf{Fitted Q iteration (FQI)} \citep{gordon1999approximate, riedmiller2005neural} assumes that the entire learning dataset of agent interactions is available from the start. If $\mathcal{D}$ represents the dataset consisting of $(s, a, r, s')$, and $w$ represents the weights of the function approximator, then the  problem can be formulated as a supervised learning regression problem by minimizing the following:
\begin{equation} \label{eq:fitted loss}
    L(w) = \frac{1}{|\mathcal{D}|}\sum_{(s, a, r, s') \in \mathcal{D}} \left[ \left( y - Q(s, a; w)\right)^2\right],
\end{equation}
where $y = r + \gamma \max_{a \in \A}Q(s', a; w)$.

\textbf{Deep Q-Networks (DQN)} \citep{mnih2015human} incorporate a deep neural network, parameterized by $w$, as a function approximator for the action-value function of the optimal policy. The neural network parameters are estimated by minimizing the squared temporal difference residual:
\begin{equation}
    \label{eq:dqn}
    L(w) = \mathbb{E}_{\mathcal{D}}\left[ \left( y- Q(s, a; w)\right)^2\right],
\end{equation}
where $y = r + \gamma \max_{a \in \A}Q(s', a; w^{-})$ and $\mathbb{E}_{\mathcal{D}}$ denotes the expectation over transitions $(s, a, r = r(s, a), s' \sim p(s' \cbar s, a))$ sampled uniformly from a replay buffer $\mathcal{D}$ of recent observed transitions. Here $w^{-}$ denotes the parameters of a target network which is updated ($w^{-} \leftarrow w$) regularly and held fixed between individual updates of $w$.
The action-value function defines the policy implicitly by $\pi(s) = \arg \max_{a \in \A}Q(s, a)$. 

\textbf{Double DQN (DDQN)} \citep{hasselt2018double} improves DQN by removing its over-estimation bias by simply replacing $y$ in equation~{\ref{eq:dqn}} with the following:
\begin{equation*}
    y = r + \gamma Q(s', a; w^{-}), \quad a = \arg \max_{a' \in \A} Q(s', a'; w).
\end{equation*}

\textbf{Deep Deterministic Policy Gradients (DDPG)} \citep{lillicrap2015continuous} is an off-policy actor-critic \citep{sutton2000policy} algorithm that uses the deterministic
policy gradient to operate over continuous action spaces. Similar to DQN, the critic estimates the value function by minimizing the same loss as in equation~{\ref{eq:dqn}} but by replacing the target $y$ with the following:
\begin{equation}
    \label{eq:ddpg}
    y =  r + \gamma Q(s', \pi(s' ; \psi); w^{-}),
\end{equation}
where $\pi(. ; \psi)$ is the parameterized actor or policy. The actor is trained to maximize the critic’s estimated value function by
back-propagating through both networks. The exploration policy in DDPG is attained by adding noise, $\pi(s; \psi) + \epsilon$, where $\epsilon \sim OU(0, \sigma^2)$ and $OU$ denotes the Ornstein-Uhlenbeck process \citep{uhlenbeck1930theory}. 

\section{VARIATIONAL INFERENCE FOR BAYESIAN NEURAL NETWORKS}
\label{sec:VI for BNN}

In order to explore more efficiently, our approach captures the uncertainty of the value estimates by using Bayesian inference. Instead of maintaining a point estimate of the deep Q-network parameters, we infer a posterior distribution. However, due to the nonlinear aspect of neural networks, obtaining the posterior distributions is not tractable and approximations have to be introduced. Thus, in this work, we use the variational inference procedure~\citep{hinton93keeping} and the so-called reparametrization trick for neural networks \citep{kingma2013auto, rezende2014stochastic}.

\textbf{Variational Inference.} Let $\mathcal{D}$ be a dataset consisting of input output pairs $\{ (x_1, y_1), \ldots (x_n, y_n) \}$. A neural network parameterized by weights $w$ models the conditional probability $p(y \cbar x, w)$ of an output $y$ given an input $x$. Let $p(w)$ and $q_{\phi}(w)$ be respectively the prior and approximate posterior over weights $w$. Variational Inference (VI) consists of maximizing the following Evidence Lower Bound (ELBO) with respect to the variational posterior parameters $\phi$:
\begin{equation}
\mathcal{L}(\phi) = \E_{q_{\phi}(w)}\left[ \log p(y \cbar x, w)\right] - \mathbb{KL}(q_{\phi}(w)  || p(w)).
\label{eq:ELBO}
\end{equation}
where $\E_{q_{\phi}(w)}$ denotes expectation over parameters $w$ sampled from $q_{\phi}(w)$.
We note that the ELBO is a lower bound on the marginal log-likelihood of the dataset $\mathcal{D}$.

\textbf{Mean Field Approximation.} \citet{blundell2015weight} assumes a mean field with independent Gaussian distributions for each weight: Let $w \in \mathbb{R}^{D_{in} \times D_{out}}$ be the weight matrix of a fully connected layer, $q_\phi(w) = \prod_{i=1}^{D_{in}} \prod_{j=1}^{D_{out}} q_{\phi}(w_{i,j}) $ and $ q_{\phi}(w_{i,j}) = \mathcal{N}( \mu_{i, j}, \sigma_{i,j}^2)$ where $ \phi = (\mu, \sigma)$ are learned parameters. The uni-modal and the fully factorized Gaussian are both limiting assumptions for high dimensional weights. They are not flexible enough to capture the true posterior distribution which is much more complex. Thus, the accuracy of the model's uncertainty estimates are potentially compromised.

\textbf{Multiplicative Normalizing Flows (MNF).} \citet{louizos2017multiplicative} use multiplicative noise to define a more expressive approximate posterior. Multiplicative noise is often used as stochastic regularization in training a deterministic neural network, such as Gaussian Dropout \citep{srivastava2014dropout}. The technique was later reinterpreted as an algorithm for approximate inference in Bayesian neural networks \citep{kingma2015variational}.
The approximate posterior is as follows:
\begin{equation}
    z \sim q_{\phi}(z); \quad 
    w \sim q_{\phi}(w \cbar z) = \prod_{i=1}^{D_{in}} \prod_{j=1}^{D_{out}} \mathcal{N}(z_i \mu_{i, j}, \sigma_{i,j}^2).
    \label{eq:MNF}
\end{equation}
The approximate posterior is considered an infinite mixture $q(w) = \int q(w \cbar z) q(z) dz$, where $z \in \mathbb{R}^{D_{in}}$ plays the role of an auxiliary latent variable (\citet{salimans2015markov, ranganath2016hierarchical}). The vector $z$ is of much lower dimension ($D_{in}$) than $w$ ($D_{in} \times D_{out}$). To make the posterior approximation richer and allow arbitrarily complex dependencies between the components of the weight matrix, the mixing density $q(z)$ is modeled via normalizing flows \citep{rezende2015variational}. This comes at an additional computational cost that scales linearly in the number of parameters. 

Normalizing flows is a class of invertible deterministic transformations for which the  determinant of the Jacobian can be computed efficiently. A rich density function $q(z_K)$ can be obtained by applying a invertible transformation $f_k$ on an initial random variable $z_0$, $K$ times, successively. Consider a simple distribution, factorized Gaussian, $q(z_0) = \prod_{j=1}^{D_{in}} \mathcal{N}(\mu_{z_{0,j}}, \sigma_{z_{0, j}}^2)$, the computation is then as follows:
\begin{align}
    z_K &= \text{NF}(z_0) = f_K \circ \ldots \circ f_1(z_0); \\
    \log q(z_K) &= \log q(z_0) - \sum_{k=1}^K\log \left| \text{det} \frac{\partial f_k}{\partial z_{k-1}} \right|.
\end{align}
\skip -0.1in
In \textit{multiplicative normalizing flows} (MNF) \citep{louizos2017multiplicative}, $z_k$ acts multiplicatively on the mean to the weights $w$ as shown in equation~\ref{eq:MNF}. We denote $\phi$ as the learnable posterior parameters which are composed of  $\mu_{z_0}, \sigma_{z_0}, \mu_w, \sigma_w$ and Normalizing flow ($\textrm{NF}$) parameters.

Unfortunately, the $\mathbb{KL}$ divergence term in the ELBO defined in equation~\ref{eq:ELBO} becomes generally intractable as the posterior $q(w)$ is an infinite mixture. This is addressed by introducing an auxiliary posterior distribution $r_{\theta}(z \cbar w)$ parameterized by $\theta$ and using it to further lower bound the $\mathbb{KL}$ divergence term of equation~\ref{eq:ELBO}. Formally, $r_{\theta}(z \cbar w)$ is parameterized by inverse normalizing flows as follows:
\begin{align}
    z_0 &\sim r(z_0 \cbar w) = \prod_{i=1}^{D_{in}} \mathcal{N}(\tilde{\mu}_i(w), \tilde{\sigma}_i(w)), \label{eq: aux posterior} \\
    \text{where} &\ \text{$\tilde{\mu}_i(w)$ and $\tilde{\sigma}_i(w)$ can be functions of $w$} \nonumber 
\end{align}
\begin{align}    
    z_0 &= \text{NF}^{-1}(z_K) = g_1^{-1} \circ \ldots \circ g_K^{-1}(z_K); \\
    \log r(z_k \cbar w) &= \log r(z_0 \cbar w) + \sum_{k=1}^K\log \left| \text{det} \frac{\partial g_k^{-1}}{\partial z_{k}} \right| ,
\end{align}

and where we parameterize $g_1^{-1}, \ldots, g_K^{-1}$ as another normalizing flow. The parameters $\theta$ are the learnable auxiliary network parameters which are composed of the parameters of $\tilde{\mu}$ and $\tilde{\sigma}$ and the parameters of the inverse normalizing flows $\textrm{NF}^{-1}$. Finally, we obtain the lower bound that MNF should optimize by replacing the $\mathbb{KL}$ divergence term with the lower bound in terms of the distribution $r_{\theta}$:
\begin{align}
    & \mathcal{L}(\phi, \theta) = \E_{q_{\phi}(w, z_K)}[ \log p(y \cbar x, w) - \label{eq:MNFELBO} \\ & \underbrace{(\mathbb{KL}(q_{\phi}(w \cbar z_K) || p(w)) - \log r_{\theta}(z_K | w) + \log q_{\phi}(z_K))}_{\textrm{regularization\_cost}(w, z_k, \phi, \theta)} ]. \nonumber
\end{align}

We can now parameterize the random sampling from $q_\phi(w, z_K)$ in terms of noise variables $\epsilon_w$ and $\epsilon_z$, and deterministic function $h$ by $w, z_K = h(\phi, \epsilon_w, \epsilon_z)$ as described by the following sampling procedure:
\begin{align}
    \epsilon_z &\sim q(\epsilon_z) = \mathcal{N}(0, \mathbf{I}_{D_{in}}), \nonumber \\
    z_0 &= \mu_{z_0} + \sigma_{z_0} \odot \epsilon_z, \nonumber\\ z_K &= \textrm{NF}(z_0) \nonumber, \\
    \epsilon_w &\sim q(\epsilon_w) = \mathcal{N}(0, \mathbf{I}_{D_{in}\times D_{out}}), \nonumber \\
    w &= (z_K \mathbf{1}_{D_{out}}^{\top}) \odot \mu_w + \sigma_w \odot \epsilon_w,
    \label{eq:MNF sampling}
\end{align}
where $\odot$ is elementwise multiplication, $\mathbf{I}_{D_{in}}$ and $\mathbf{I}_{D_{in}\times D_{out}}$ are identity matrices and $\mathbf{1}_{D_{out}}$ is a $D_{out}$ dimensional vector whose entries are all equal to one. The lower bound $\mathcal{L}(\phi, \theta)$ in equation \ref{eq:MNFELBO} can be written as:
\begin{align}
    & \mathcal{L}(\phi, \theta) = \E_{\epsilon_z, \epsilon_w}[ \log p(y \cbar x, h(\phi, \epsilon_w, \epsilon_z)) - \nonumber \\
    & \quad \textrm{regularization\_cost}(h(\phi, \epsilon_w, \epsilon_z), \theta)]
\end{align}
Thus, we can have a Monte Carlo sample of the gradient of $\mathcal{L}(\phi, \theta)$ with respect to $\phi$ and $\theta$. This parameterization allows us to handle the approximate parameter posterior as a straightforward optimization problem. 

\textbf{Choice of normalizing flows.} 
In practice, we use masked RealNVP \citep{dinh2016density} for the normalizing flows. In particular, we use the numerically stable updates described in Inverse Autoregressive Flow \citep{kingma2016improved}:
\begin{align*}
m = \textrm{Bern}(0.5), &\quad h = \tanh(a(m \odot z_k)), \\ 
\mu = b(h),  &\quad \sigma = \textrm{sigmoid}(c(h)),
\end{align*}
\begin{align}
    z_{k+1} &= f_k(z_k)  \\
    &= m \odot z_k + \nonumber \\ 
    & \quad (1-m)\odot(z_k \odot \sigma + (1-\sigma) \odot \mu), \nonumber \\
    \log \left| \frac{\partial f_k}{\partial z_k}\right| &= (1-m)^{\top}\log \sigma ,
\end{align}

where $a(.), b(.)$ and $c(.)$ are linear transformations. For the auxiliary posterior distribution $r(z_0 \cbar w)$ defined in equation \ref{eq: aux posterior}, we parameterize the mean and the standard deviation $\tilde{\mu}$ $\tilde{\sigma}$ as in the original paper \citep{louizos2017multiplicative}.


\section{MULTIPLICATIVE NORMALIZING FLOWS FOR RANDOMIZED VALUE FUNCTIONS}
\label{sec: MNF for RVF}
 

We now turn to our novel proposed approach that incorporates the techniques we previously introduced to modify both DQN and DDPG in a similar fashion. In particular, we introduce a new parametrization of the value functions (optimal value function in DQN and the critic in DDPG) and change their corresponding losses.

We model the distribution of target return $y$ ($y$ is equal to $r + \gamma \max_{a \in \A}Q(s', a; w^{-})$ for DQN and to $r + \gamma Q(s', \pi(s'; \psi); w^{-})$ for DDPG) as a Gaussian distribution with parameterized mean $Q(s, a; w)$ \footnote{We overload our notation $w$ for both the weight matrix of a single layer and the full set of network parameters.} and constant standard deviation $\tau$: 
$ y \sim p(y \mid s, a) = \mathcal{N}(Q(s, a; w), \tau^2) $.

In this setting, minimizing the loss of DQN or of the DDPG critic corresponds to a maximum log-likelihood estimation. Indeed $L(w)$ from equation \ref{eq:fitted loss} is such that:
\begin{align}
  | \mathcal{D}| L(w) &= \sum_{(s, a, r, s') \in \mathcal{D}}\left[ \left( y- Q(s, a; w)\right)^2\right] \nonumber \\
  &= 
    - 2 \tau^2 \sum_{(s, a, r, s') \in \mathcal{D}} \log p( y \cbar s, a),
\end{align}
where we ignore constant terms. Instead of computing a maximum likelihood estimate of the parameters $w$ of the value function network, we use a randomized value function to track an \emph{approximate posterior distribution} over  network parameters $w$, using the MNF family to parameterize this  posterior. The weights $w$ are considered random variables and are obtained by the sampling procedure described in equation \ref{eq:MNF sampling}. The value function $Q(s,a, \epsilon_z, \epsilon_w; \phi, \theta)$ is now parameterized in terms of the approximate posterior $(\phi, \theta)$ defined in Section~\ref{sec:VI for BNN}. Our approach optimizes the ELBO in equation~\ref{eq:MNFELBO} with respect to the approximate posterior parameters $(\phi, \theta)$, which amounts to minimizing the following loss:
\begin{align}
  & \E_{\epsilon_z, \epsilon_w} [\sum_{(s, a, r, s') \in \mathcal{D}} \left( y - Q(s, a, \epsilon_z, \epsilon_w; \phi, \theta)\right)^2 + \nonumber \\
  & \quad 2 \tau^2  \textrm{regularization\_cost}(\epsilon_z, \epsilon_w; \phi, \theta) ] ,
    \label{eq:MNFDQN loss}
\end{align}
where $y =  r + \gamma \max_{a}Q(s, a, 0, 0; \phi^{-}, \theta^{-})$ for DQN, $y = \gamma Q(s, \pi(s'; \psi), 0, 0; \phi^{-}, \theta^{-})$ for DDPG and where the noise is disabled for the target network.  
This loss is amenable to mini-batch optimization. In a supervised learning setting, for each mini-batch $\mathcal{M} \subset \mathcal{D}$, we would take a gradient step to lower the following loss:
\begin{align}
    &\E_{\epsilon_z, \epsilon_w} [ \frac{1}{| \mathcal{M}|} \sum_{(s, a, r, s') \in \mathcal{M}} 
    \left( y - Q(s, a, \epsilon_z, \epsilon_w; \phi, \theta)\right)^2 + \nonumber\\
    &\quad \lambda  \; \textrm{regularization\_cost}(\epsilon_z, \epsilon_w; \phi, \theta) ],
    \label{eq: minibatch loss}
\end{align}
where $\lambda = \frac{2 \tau^2}{|\mathcal{D}|}$. This makes the regularization cost uniformly distributed among mini-batches at each epoch. In the RL setting, however, we only keep a moving window of experiences in the replay buffer. Thus, the size of replay buffer $|\mathcal{D}|$ is not directly analogous to the size of the dataset. So we leave $\lambda$ as a hyper-parameter to tune.

The expectation in equation \ref{eq: minibatch loss} is with respect to the distribution of noise variables $\epsilon_z$ and $\epsilon_w$ of the online value function. An unbiased estimate of the loss is thus obtained by simply sampling the two noise variables. The current noise samples are held fixed across the mini-batch. The learnable parameters $(\phi, \theta)$ are then updated by performing a gradient descent on the mini-batch loss. In the case of DDPG, in addition to the critic parameters, we update the actor policy parameters $\psi$ using the following sampled policy gradient: 
\begin{equation*}
    \frac{1}{|\mathcal{M}|} \sum_{s \in \mathcal{M}} \nabla_{a} Q(s, a, \epsilon_z, \epsilon_w; \phi, \theta)|_{a = \pi(s; \psi)} \nabla_{\psi} \pi(s; \psi)
\end{equation*}
After the update, we generate new noise samples and we select actions greedily with respect to the corresponding value function.

We call the new adaptations of DQN and DDPG, MNF-DQN and MNF-DDPG, respectively. Detailed algorithms are provided as Algorithm~\ref{alg:mnf} and Algorithm~\ref{alg:mnf-ddpg} (MNF-DDPG could be find in appendix).

\begin{algorithm}[H] 
  \caption{MNF-DQN}
    \label{alg:mnf}  
  \begin{algorithmic}[1]
  \State \textbf{Input:} $m$ mini-batch size; $D$ empty replay buffer; $K$ update frequency, $U$ target update
  frequency
  \For{episode $e \in {1,  ..., M}$}  
     \State Set $s \leftarrow s_0$ 
     \State Sample noise variables $\epsilon_w$ and $\epsilon_z$ from standard normal distribution.
     \For{$t \in {1, \dots}$}
        \State Select an action $a_t \leftarrow \arg \max_{a\in A} Q(s, a, \epsilon_z, \epsilon_w; \phi, \theta)$
        \State Observe next state $s_{t+1}$ and reward ${r_t}$ after taking action $a_t$
        \State Store transition $(s_t, a_t, r_t, s_{t+1})$ in the replay buffer $D$
        \If{$t \mod K == 0$}
            \State Sample a mini-batch of $m$ transitions $((s_j, a_j, r_j, s_{j}') \sim D)^m_{j=1}$
            \For{$j=1, \ldots, m$}
                \If{ $s_j'$ is terminal state}
                    $y_j = r_j$ 
                \Else
                    \State $y_j = r_j + \gamma Q(s, a, 0, 0; \phi^{-}, \theta^{-})$
                \EndIf
            \EndFor 
            \State Re-sample noise variables $\epsilon_w$ and $\epsilon_z$ and perform gradient step with respect to $(\phi, \theta)$:
            \State $\frac{1}{m} \sum_{j=1, \ldots m} 
    \left( y_j - Q(s_j, a_j, \epsilon_z, \epsilon_w; \phi, \theta)\right)^2 + \lambda    \textrm{regularization\_cost}(\epsilon_z, \epsilon_w; \phi, \theta)$
    \State Re-sample noise variables $\epsilon_w$ and $\epsilon_z$
        \EndIf
        \If{$t \mod U == 0$}
            \State Set $\theta^{-}, \phi^{-} \leftarrow \theta, \phi$
        \EndIf
        \State Set $s \leftarrow s_{t+1}$
     \EndFor
  \EndFor
  
  \end{algorithmic} 
 
\end{algorithm}

\section{RELATED WORK}

There have been several recent works that incorporate Bayesian parameter updates with deep reinforcement learning for efficient exploration.

\citet{osband2016deep} propose BootstrappedDQN which consists of a simple non-parametric bootstrap with random initialization to approximate a distribution over value functions. BootstrappedDQN consists of a network with multiple Q-heads. At the start of each episode, the agent samples a head, which it follows for the duration of the episode. BootstrappedDQN is a non-parametric approach to uncertainty estimation. In contrast, MNF-DQN uses a parametric approach, based on variational inference, to quantify the uncertainty estimates.

\citet{azizzadenesheli2018efficient} extend randomized least-square value iteration by \citet{osband2016generalization} (which was restricted to linear approximator) to deep neural networks. In particular, they consider only the last layer as stochastic and keep the remaining layers deterministic. As the last layer is linear, they propose a Bayesian linear regression to update the posterior distribution of its weights in closed form. In contrast, our method is capable of performing an approximate Bayesian update on the full network parameters. Variational inference could be applied for all layers using stochastic gradient descent on the approximate posterior parameters. 

The closest work to ours is BBQ-Networks by \citet{lipton2016efficient}. Their algorithm, called Bayes-by-Backprop Q-Network (BBQN), uses variational inference to quantify uncertainty. It uses independent factorized Gaussians as an approximate posterior \citep{blundell2015weight}. In our work, we argue that to achieve efficient exploration, we need to capture the true uncertainty of the value function. The latter depends importantly on the flexibility of the approximate posterior distribution. We also note that BBQN was proposed for Task-Oriented Dialogue Systems and was not evaluated on standard RL benchmarks.
Furthermore, BBQN can be seen simply as a sub-case of MNF-DQN. In fact, the two algorithms are equivalent when we set the auxiliary variable $z$ in MNF-DQN to be equal to one.

Our work is also related to methods that inject noise in the parameter space for exploration. For such methods, at the beginning of each episode, the parameters of the current policy are perturbed with some noise. This results in a policy that is perturbed but still consistent for individual states within an episode. This is sometimes called state-dependent exploration~\citep{sehnke2010parameter} as the same action will be taken every time the same state is sampled in the episode. Recently, \citet{fortunato2017noisy} proposed to add parametric noise to the parameters of a neural network and show that its aids exploration. The parameters of the noise are learned with gradient descent along with the remaining network weights. Concurrently, \citet{plappert2017parameter} proposed a similar approach but they rely on heuristics to adapt the noise scale instead of learning it as in \citet{fortunato2017noisy}.  

\citet{osband2018randomized} recently discuss limitations of some popular approaches for exploration in Deep RL. In particular, approaches such as NoisyDQN, BBQ-Networks and even our proposed method MNF-DQN treat each Bellman error as an independent draw from the posterior. This fact, according to \citet{osband2018randomized}, prevents these methods from propagating uncertainty in the correct way. On the other hand, BootstrapedDQN would be able to propagate a temporally-consistent sample of Q-values, since each head is trained only on its own target value. However, we argue that MNF-DQN is still a principled approach: 
We can view DQN from the value iteration perspective. DQN considers a target network $\tilde{Q}$ and tries to minimize the mean-squared Bellman error (MSBE) $|| Q - \mathcal{T} \tilde{Q}||^2$ where $\mathcal{T}$ is the optimality Bellman operator. For a given target network, DQN tries to solve the latter regression problem by performing multiple stochastic gradient steps. This can be viewed as one-step of value iteration that tries to fit $Q$ to $\mathcal{T} \tilde{Q}$. Maintaining only a point estimate of $Q$ does not capture the prediction uncertainty and leads to decisions that do not reflect our understanding of the environment. Therefore, instead of having a point estimate of $Q$, we turn the mean-squared Bellman error minimization into bayesian regression and maintain a posterior distribution over $Q$. This aims at capturing the parametric uncertainty due to limited data and prevents overfitting. When $(s, a)$ are poorly understood, the distribution of $Q(s, a)$ might have large variance. When we draw a sample from this distribution, there is a chance that we attribute a high value to $(s, a)$ due to the variance which in turn enables exploration. As long as we collect more data, the variance should decrease, which enable exploitation.



\section{EXPERIMENTS}
We evaluate the performance of MNF-DQN on two toy domains (N-chain and Mountain Car), as well as on several Atari games \citep{bellemare2013arcade}. Moreover, we compare MNF-DDPG on continuous control tasks from OpenAI Gym \citep{brockman2016openai}. We compare the performance of our agents to several recent state-of-the-art deep exploration methods.

\subsection{DISCRETE ACTION ENVIRONMENTS}
\subsubsection{Toy Tasks}

\paragraph{$n$-Chain} As a sanity check, we evaluate MNF-DQN on the well-known n-chain environment introduced in \citet{osband2016deep}. The environment consists of N states. The agent always starts at the second state $s_2$ and has two possible actions: move right and move left. A small reward $r=0.001$ is received in the first state $s_1$, a large reward $r=1$ in the final state $s_N$, otherwise the reward is zero. 

\begin{figure*}
\centering
\includegraphics[width=1\linewidth]{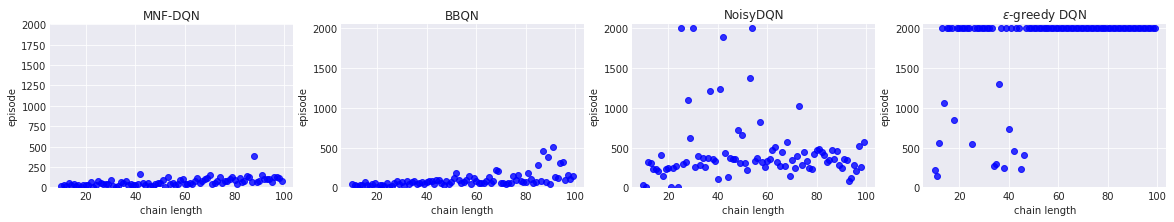}
\caption{\label{fig:nchain} Median number of episodes (max 2000) required to solve the n-chain problem for (figure from left to right) MNF-DQN, BBQN, NoisyDQN and $\epsilon$-greedy DQN. The median is obtained over 10 runs with different seeds. We see that MNF-DQN consistently performs best across different chain lengths.}
 \vskip -0.1in
\end{figure*}

We compare the exploration behavior of MNF-DQN, NoisyDQN~\citep{fortunato2017noisy}, BBQN~\citep{lipton2016efficient} and $\epsilon$-greedy DQN on varying chain lengths. We train each agent with ten different random seeds for each chain length. After each episode, agents are evaluated on a single roll-out with all of their randomness disabled ($\epsilon$ is set to zero for DQN, noise variables are set to zero for MNF-DQN, BBQN, NoisyDQN). The problem is considered solved when the agent completes the task optimally for one hundred consecutive episodes. While the task is admittedly a simple one, it still requires adequate exploration in order to be solved. This is especially true with large chain lengths, as it is easy to discover the small reward and fall into premature exploitation. 
Figure \ref{fig:nchain} shows that MNF-DQN has very consistent performance across different chain lengths. MNF-DQN clearly outperforms $\epsilon$-greedy DQN which most of the time fails to solve the problem when $n \geq 10$. BBQN performs well but slightly worse than MNF-DQN for very large chain length. MNF-DQN also outperforms NoisyDQN, which on average needs a larger number of episodes to solve the task.

\paragraph{Mountain Car} ~\citep{moore1990efficient} is a classic RL continuous state task where the agent (car) is initially stuck in a valley and the goal is to drive up the mountain on the right. The only way to succeed is to drive back and forth to build up momentum. We use the implementation provided by OpenAI gym~\citep{brockman2016openai}, where the agent gets $-1.0$ reward at every time-step and get $+1.0$ reward when it reaches up the mountain, at which point the episode ends. The maximum length of an episode is set to 1000 time-steps.

\begin{figure}
\centering
\includegraphics[width=0.8\linewidth]{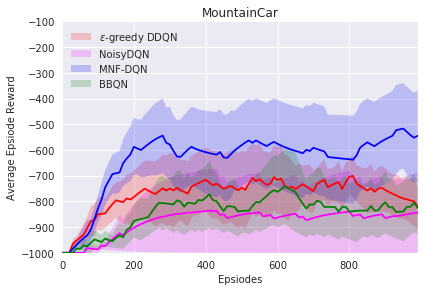}
\caption{\label{fig:mcar} Average return per episode over 5 runs in MountainCar.}
\vskip -0.1in
\end{figure}

We evaluate the performance of the agent every 10 episodes by using no noise over 5 runs with different random seeds. As can be seen in  Figure~\ref{fig:mcar}, MNF-DQN learns much faster and performs better when compared to the other exploration strategies.

\subsubsection{Arcade Learning Environment}
\label{sec:atari-experiments}

Next, we consider a set of Atari games \citep{bellemare2013arcade} as a benchmark for high dimensional state-spaces. We compare MNF-DQN to standard DQN with $\epsilon$-greedy exploration, BBQN and NoisyDQN. We use the same network architecture for all agents, i.e three convolutional layers and two fully connected layers. For DQN, we linearly anneal $\epsilon$ from 1.0 to 0.1 over the first 1 million time-steps. For NoisyDQN, the fully connected layers are parameterized as noisy layers and use factorized Gaussian noise as explained in \citet{fortunato2017noisy}. For MNF-DQN and BBQ, in order to reduce computational overhead, we choose to consider only the parameters of the fully connected layers as stochastic variables and perform variational inference on them. We consider the parameters of the convolutional layers as deterministic variables and optimize them using maximum log-likelihood. For MNF-DQN, the normalizing flows are of length two for $q_\phi$ and $r_\theta$, with 50 hidden units for each step of the flow. To have a fair comparison across all algorithms, we fill the replay buffer with actions selected at random for the first 50 thousand time-steps.

\begin{figure*}
\centering
\includegraphics[width=0.8\linewidth]{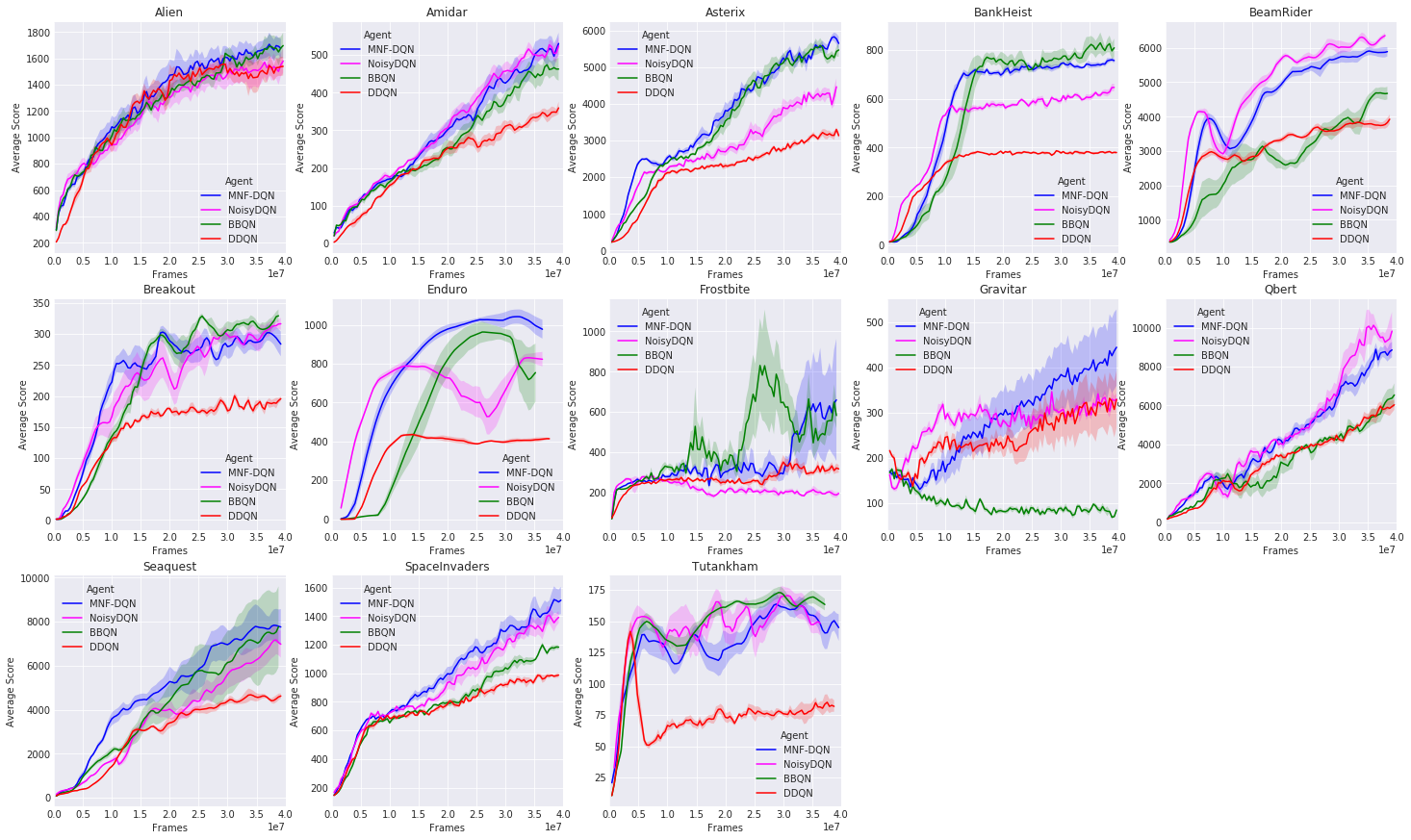}
\caption{\label{fig:atari} Comparison of the training curves of MNF-DQN agent versus the NoisyDQN, BBQN and DQN + $\epsilon$-greedy baselines for various Atari games. }
\vspace*{-5mm}
\end{figure*}

We use the standard hyper-parameters of DQN for all agents. MNF-DQN and BBQN have an extra hyper-parameter $\lambda$, the trade-off parameter between the log likelihood cost and the regularization cost. To tune $\lambda$, we run MNF-DQN and BBQN for $\lambda = \frac{\alpha}{|\mathcal{D}|}$ where $\alpha \in \{10^{-6}, 10^{-5}, 10^{-4}, 10^{-3}, 10^{-2} \}$. As explained earlier, $|\mathcal{D}|$ is not good proxy for dataset size but it gives a good value range. We still tune the hyperparameter $\alpha$.
We train each agent for 40 millions frames. We evaluate each agent on the return collected by the exploratory policy during training steps. Each agent is trained for 5 different random seeds. We plot in Figure~\ref{fig:atari} the median return as well as the interquartile range.

From Figure~\ref{fig:atari} we see that across all games our approach provides competitive and consistent results. Moreover, the naive epsilon-greedy approach (DDQN) performs significantly worse than the exploration-based methods in most cases. MNF-DQN provides a boost in performance over the baselines in Gravitar, which is considered a \textit{hard exploration} and sparse reward game \citep{bellemare2016unifying}. BBQN fails completely for this game. In all the other \textit{hard exploration} games (Amidar, Alien, Bank Heist, and Qbert), the difference between MNF-DQN and the best performing baseline is minimal (within the margin of error). 

Note that MNF-DQN, as well as the baseline algorithms, fail to solve extremely challenging games such as 
Montezuma’s Revenge. 
Note also that methods from the literature that achieved some degree of success on Montezuma's Revenge either include prior information about the game \citep{le2018hierarchical, aytar2018playing} or are combined with heuristics on the intrinsic reward \citep{bellemare2016unifying}. The first category of methods use human demonstrations to drive the exploration, and the latter require maintaining a separate density model over the state-action space.

\begin{figure*}
\centering
\includegraphics[width=1\linewidth]{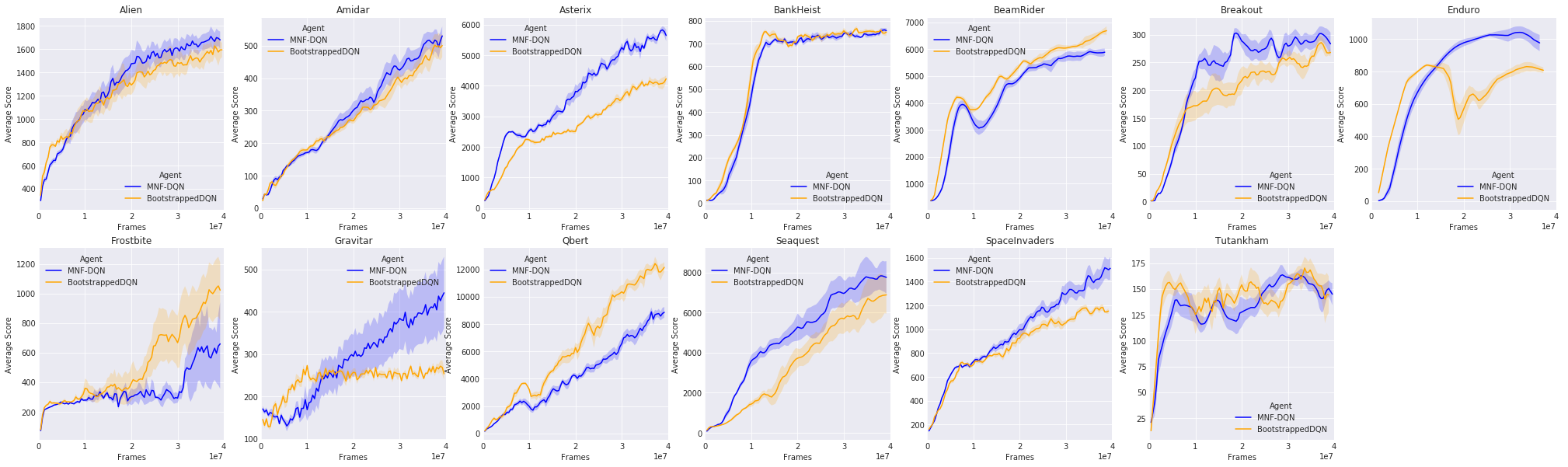}
\caption{\label{fig:bootdqn} Comparison of the training curves of MNF-DQN agent versus the Bootstrapped DQN for various Atari games. }
\vspace*{-5mm}
\end{figure*}

So far, we considered only parameterized noise baselines. Now, we compare MNF-DQN to Bootstrapped DQN \citep{osband2016deep}. Results are given in Figure \ref{fig:bootdqn}. MNF-DQN often outperforms Bootstrapped DQN (Out of 13 games, MNF-DQN clearly outperforms Bootstrapped DQN on 6 of them, yields similar performance on 4, and under-performs it only on 3). Moreover, we investigate the randomized prior function introduced by \citet{osband2018randomized} which consists in adding a randomized un-trainable prior network to each Q-head. We find that the Bootstrapped DQN with randomized prior does not improve performance over Bootstrapped DQN. 

\subsection{CONTINUOUS ACTION ENVIRONMENTS}
\label{sec:mujoco-experiments}
In this section, we benchmark MNF-DDPG on the continuous control tasks from OpenAI Gym \citep{brockman2016openai}. 
We compare our algorithm to the standard DDPG with OU-noise with $\sigma = 0.2$ (OU-noise DDPG), DDPG with BBQN based critic (BBQN-DDPG) and with Noisy network based critic (Noisy-DDPG). The actor is deterministic in all the agents and only the critic has stochasticity. We use the same setup as \citet{plappert2017parameter}, where both actor and critic use 2 hidden layers with 64 units and a ReLU non-linearity. More details about the architecture can be found in Appendix~\ref{app:mujoco-details}. 

For MNF-DDPG, the normalizing flows are of length one for $q_\phi$ and $r_\theta$, with 16 hidden units for each step of the flow. We tune the hyper-parameter $\lambda$ in manner similar to what we described  in Sec~\ref{sec:atari-experiments}, trying out $\lambda = \frac{\alpha}{|\mathcal{D}|}$ where $\alpha \in \{1e^{-6}, 4e^{-6}, 8e^{-6}, 1e^{-5}, \dots, 8e^{-2}, 1e^{-1}, 4e^{-1}, 8e^{-1} \}$. We evaluate the algorithms on 3 different control tasks, where each agent is trained for 1 million time-steps. Each epoch consists of 10,000 time-steps and after every epoch the agent is evaluated with all their randomness disabled for 20 episodes. Each agent is trained for 5 different random seeds and the results are averaged.

\begin{figure*}
\centering
\includegraphics[width=\linewidth]{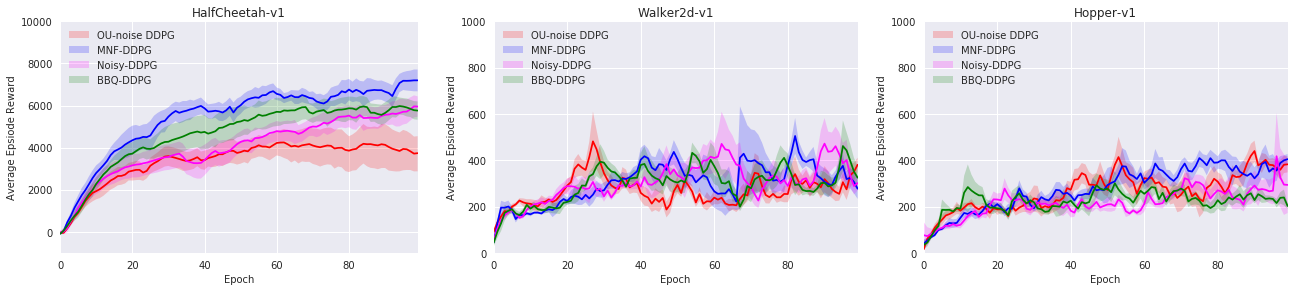}
\caption{\label{fig:mujoco} Average return per episode for continuous control environments plotted over epochs over 5 runs.}
\vskip -0.1in
\end{figure*}

Results are shown Figure~\ref{fig:mujoco}. The MNF-DDPG agent performs better than all the other baselines in the HalfCheetah environment. In the other environments there is not much difference between the different exploration methods. This finding is consistent with the results from \cite{plappert2017parameter}, where they also conclude that the other environments do not require a lot of exploration due to their well-structured reward function.

\section{CONCLUSION}
\vskip -0.1in

Through the combination of multiplicative normalizing flows and modern value-based deep reinforcement learning methods, we show that a powerful approximate posterior can be efficiently utilized for better exploration. Moreover, the improved sample efficiency comes only at a computational cost that is linear in the number of model parameters. Finally, we find that on several common Deep RL benchmarks, the MNF approximation outperforms state-of-the-art exploration baselines.

\clearpage

\bibliographystyle{apalike}
\bibliography{lib}


\clearpage

\onecolumn  

\appendix

\section{Derivation of the lower bound}
We have the following ELBO and we would like to lower bound the KL term using the auxiliary posterior distribution $r_\theta(z \mid \omega)$
\begin{equation}
\mathcal{L}(\phi) = \E_{q_{\phi}(w)}\left[ \log p(y \cbar x, w)\right] - \mathbb{KL}(q_{\phi}(w)  || p(w)).
\label{eq:ELBO}
\end{equation}

As the KL divergence is always non-negative, we have:
\begin{align}
    \mathbb{KL}(q_{\phi}(w)  || p(w)) & \leq \mathbb{KL}(q_{\phi}(w)  || p(w)) + \E_{q_{\phi}(\omega)}\left [\mathbb{KL}(q_{\phi}(z | w)  || r_\theta(z | w)) \right] \\
    & = \E_{q_\phi(\omega)}\left[ \log q_\phi(\omega) - \log p(\omega) \right] + \E_{q_\phi(\omega, z)} \left[ \log q_\phi(z \mid \omega) - \log r_\theta (z \mid \omega)\right] \\
    & = \E_{q_\phi(\omega, z)} \left[ \log ( q_\phi(\omega) q_\phi(z \mid \omega) ) - \log p(\omega) - \log r_\theta(z \mid \omega)\right] \\
    & = \E_{q_\phi(\omega, z)} \left[ \log ( q_\phi(\omega \mid z) q_\phi(\omega) ) - \log p(\omega) - \log r_\theta(z \mid \omega)\right] \\
    & = \E_{q_\phi(\omega, z)} \left[ \mathbb{KL}(q_\phi(\omega \mid z) || p(\omega))
    + \log q_\phi(z) - \log r_\theta(z \mid \omega \right]
\end{align}

which proves the lower bound in equation \ref{eq:MNFELBO}

\section{Algorithms}
\label{sec:appx_algo}

\begin{algorithm} 
  \caption{MNF-DDPG algorithm}
    \label{alg:mnf-ddpg}  
    \begin{algorithmic}
    \State \textbf{Input:} $m$ mini-batch size; $D$ empty replay buffer; $K$ update frequency, $U$ target update frequency
    \State Initialize critic network $ Q(s, a, \epsilon_z, \epsilon_w ; \phi, \theta)$ and actor $\pi(s;  \psi)$ with weights $ \theta$, $\phi$ and $ \psi$. 
    \State Initialize target critic network $Q(s,a, \epsilon_z, \epsilon_w; \phi^{-1}, \theta^{-})$ and actor $\pi(s; \psi^{-})$ with weights $\theta^{-} \leftarrow \theta, \phi^{-1} \leftarrow \phi,  \psi^{-} \leftarrow \psi$
    \For{episode $e \in {1,  ..., M}$}  
     \State Set $s \leftarrow s_0$ 
     \State Sample noise variables $\epsilon_w$ and $\epsilon_z$ from standard normal distribution.
     \For{$t \in {1, \dots}$}
        \State Select an action $a_t \leftarrow \pi(s; \psi)$
        \State Observe next state $s_{t+1}$ and reward ${r_t}$ after taking action $a_t$
        \State Store transition $(s_t, a_t, r_t, s_{t+1})$ in the replay buffer $D$
        \If{$t \mod K == 0$}
            \State Sample a mini-batch of $m$ transitions $((s_j, a_j, r_j, s_{j}') \sim D)^m_{j=1}$
            \For{$j=1, \ldots, m$}
                \If{ $s_j'$ is terminal state}
                    $y_j = r_j$ 
                \Else
                    \State $y_j = r_j + \gamma Q(s, \pi(s; \psi^{-}), 0, 0; \phi^{-1} \theta^{-})$
                \EndIf
            \EndFor 
            \State Re-sample noise variables $\epsilon_w$ and $\epsilon_z$ and perform gradient step with respect to $(\phi, \theta)$:
            \State $\frac{1}{m} \sum_{j=1, \ldots m} 
    \left( y_j - Q(s_j, a_j, \epsilon_z, \epsilon_w; \phi, \theta)\right)^2 + \lambda   \textrm{regularization\_cost}(\epsilon_z, \epsilon_w; \phi, \theta)$

    \State Update the actor using sampled policy gradient: $\frac{1}{m} \sum_{j=1, \ldots m} \nabla_{a} Q(s_j, a, \epsilon_z, \epsilon_w; \phi, \theta)|_{a = \pi(s_j; \psi)} \nabla_{\psi} \pi(s_j; \psi)
    $
    
    \State Re-sample noise variables $\epsilon_w$ and $\epsilon_z$

        \EndIf
        \If{$t \mod U == 0$}
            \State Set $\theta^{-} \leftarrow \tau \theta + (1 - \tau) \theta^{-}$
            \State Set $\phi^{-} \leftarrow \tau \phi + (1 - \tau) \phi^{-}$
            \State Set $\psi^{-} \leftarrow \tau \psi + (1 - \tau) \psi^{-}$
        \EndIf
        \State Set $s \leftarrow s_{t+1}$
     \EndFor
  \EndFor
  
  \end{algorithmic} 
 
\end{algorithm}

\section{Continuous Control Architecture Details}
\label{app:mujoco-details}

The actor and critic use 2 hidden layers each with 64 ReLU units. The action is added to the second hidden layer of the critic. Layer-normalization is applied between the hidden layers before the non-linearity and target-networks are soft updated with $\tau=0.001$. Critic is trained with learning rate = $10^{-3}$ and the actor with $10^{-4}$. Adam optimizer is used with batch size of 128 to update the weights. Replay buffer has size of $100K$ transitions with $\gamma=0.99$. Both update frequency ($K$) and target update frequency ($U$) are set to 1 (in Alg.\ref{alg:mnf-ddpg}).

\section{Comparison with Bayesian DQN}
We did not implement Bayesian DQN \citep{azizzadenesheli2018efficient} but we used the results of 14 Atari games provided by authors in the article's github repository \url{https://github.com/kazizzad/BDQN-MxNet-Gluon}. 
Out of these 14 games, seven are in common with our set of games. Among these 7 games, Bayesian DQN outperforms our method only in 3 games. See Figure~\label{fig:bayesiandqn}.

\begin{figure}[h]
\centering
\includegraphics[width=1\linewidth]{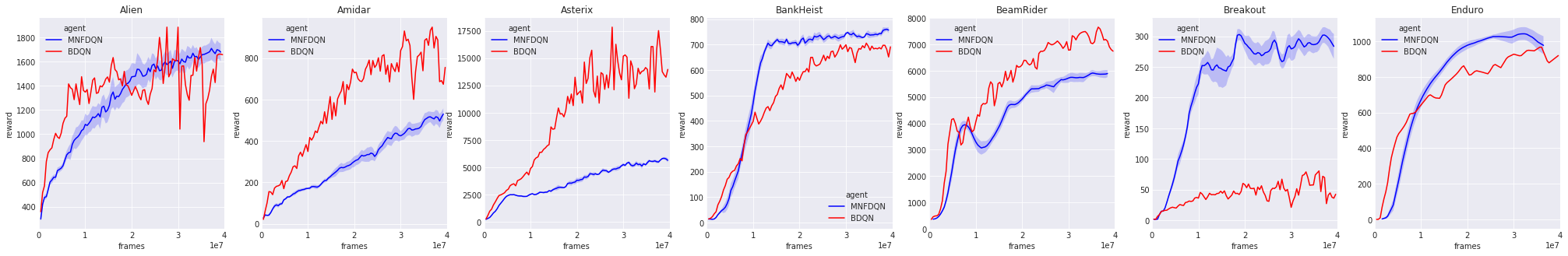}
\caption{\label{fig:bayesiandqn} Comparison of the training curves of MNF-DQN agent versus the Bayesian DQN \citep{azizzadenesheli2018efficient} for various Atari games.  }
\vspace*{-5mm}
\end{figure}


\end{document}